\documentclass{article}

\usepackage{PRIMEarxiv}

\usepackage[utf8]{inputenc} 
\usepackage[T1]{fontenc}    
\usepackage{hyperref}       
\usepackage{url}            
\usepackage{booktabs}       
\usepackage{amsfonts}       
\usepackage{amssymb}        
\usepackage{nicefrac}       
\usepackage{microtype}      
\usepackage{amsmath}        
\usepackage{lipsum}
\usepackage{fancyhdr}       
\usepackage{graphicx}       
\graphicspath{{media/}}     

\title{EgoLabel: Zero-Shot Egocentric Auto-Labeling with Cross-Modal Consistency
\thanks{\textit{\underline{Citation}}: 
\textbf{Authors. Title. Pages.... DOI:000000/11111.}} 
}

\author{
  Xiaozhi Zhang\textsuperscript{1,\textasteriskcentered}, Changjia Xu\textsuperscript{1},
  Yaoyuan Yan\textsuperscript{2}, Haoxiang Jie\textsuperscript{2},
  Bingjie Fu\textsuperscript{3,4} \\
  \textsuperscript{1}Robotics Lab, Country Garden Services \\
  \textsuperscript{2}AI Lab, Country Garden Services \\
  \textsuperscript{3}School of Computer Science and Engineering, Northeastern University, Shenyang \\
  \textsuperscript{4}Omni AI \\
  \texttt{zhangxiaozhi07@bgyfw.com} \\
}

\begin{document}
\maketitle

\begin{abstract}
Egocentric auto-labeling requires depth, camera pose, object, and human-pose estimates to agree across independently developed models. Differences in units, coordinate conventions, and scale can make individually plausible predictions geometrically inconsistent. We present \textbf{EgoLabel}, a zero-shot pipeline that combines off-the-shelf models without target-domain labeled training and provides explicit cross-modal registration and consistency diagnostics. The pipeline integrates MonST3R for geometry, LocateAnything with GMC-BoT-SORT and SAM~2 for objects, and V-HPOT for hands; FisheyeViT provides a separate whole-body compatibility demonstration. Our main hand evaluation uses H2O sensor depth. Across 20 sequences, per-hand depth anchoring reduces the mean of per-sequence median hand--depth residuals by $23\%$ relative to a single global scale, yielding a pooled median of $84.6\,\mathrm{mm}$ against a $20.5\,\mathrm{mm}$ ground-truth reference. Hand PA-MPJPE is $14.53\pm1.29\,\mathrm{mm}$, whereas absolute MPJPE is $119.1\pm31.3\,\mathrm{mm}$, revealing a substantial gap between aligned and metric accuracy. On three RGB-only sequences, camera ATE is $1.40\pm0.29\,\mathrm{cm}$ after $\mathrm{Sim}(3)$ alignment. We also identify a sliding-window implementation issue that allows sensor-initialized depth to change during optimization. Because hand anchoring and its residual share the same depth map, the residual measures internal consistency rather than independent accuracy. These results characterize both the utility and the current limits of the registration pipeline.
\end{abstract}


\section{Introduction}
Egocentric (first-person) vision supports activity understanding, hand--object interaction analysis, and embodied learning. Benchmarks such as Ego4D~\cite{grauman2022ego4d}, EPIC-KITCHENS~\cite{damen2022rescaling}, H2O~\cite{kwon2021h2o}, AssemblyHands~\cite{kwon2021assemblyhands}, HOT3D~\cite{hot3d2025}, and SceneEgo~\cite{sceneego2023} have enabled progress in these areas, but producing their annotations remains costly. Depth, camera trajectories, hands, objects, and body motion typically require separate capture or processing pipelines, adding work as the number of modalities increases. Automatically generating jointly registered annotations from a single recording could help scale the supply of first-person interaction data for embodied models and robot policies~\cite{wang2026evwm,jie2026omnivla}.

Recent foundation models provide many of the required components: DUSt3R~\cite{wang2024dust3r} and MonST3R~\cite{zhang2024monst3r} for geometry, HaMeR~\cite{pavlakos2024hamer} and V-HPOT~\cite{mucha2026towards} for hands, FisheyeViT~\cite{fisheyevit2024} for the body, and open-vocabulary detectors, trackers, and segmenters~\cite{liu2024groundingdino,locateanything2025,aharon2022botsort,ravi2024sam2} for objects. These models can be deployed without target-domain labeled training. In particular, monocular geometry models estimate depth and 6DoF camera trajectories from RGB, although their global scale remains ambiguous.

Combining these outputs requires more than running the models in sequence. Independently developed models use different units (millimeters, meters, or focal-length-normalized scales), axis conventions, and reference frames. A hand skeleton in virtual-camera units cannot be directly combined with a camera trajectory in meters; similarly, lifting object detections into a shared 3D scene requires compatible depth, poses, and intrinsics. Such mismatches can remain unnoticed because each model returns plausible predictions and is evaluated on its own task. Per-module accuracy therefore does not establish agreement \emph{between} annotation streams.

We present \textbf{EgoLabel}, a zero-shot toolkit for registering and inspecting heterogeneous egocentric annotations. Here, zero-shot means that no target-domain labeled training is performed. The pipeline converts V-HPOT outputs using the recording camera's intrinsics, anchors each hand to depth, and lifts object detections using depth and estimated camera poses. We evaluate cross-modal agreement through a hand--depth residual and an object-trajectory smoothness score, with ground-truth references, a constant-scale control, and a synthetic depth-scale response analysis. The main hand study uses sensor depth, while a smaller RGB-only study examines camera pose and hand--depth consistency. The registration procedure and diagnostics make the assumptions and failure modes of model composition explicit; their scope and dependence on shared inputs are specified in Section~\ref{sec:consistency}.

Our contributions are therefore:
\begin{itemize}
  \item \textbf{A zero-shot egocentric auto-labeling pipeline.} EgoLabel integrates MonST3R, LocateAnything with GMC-BoT-SORT and SAM~2, and V-HPOT into a pipeline for geometry, object, and hand annotations. A separate FisheyeViT demonstration examines whole-body compatibility, and a structured-captioning application illustrates downstream use (Sections~\ref{sec:methodology} and~\ref{sec:captioning}).
  \item \textbf{Explicit registration and consistency diagnostics.} We document unit conversion, coordinate alignment, and per-hand depth anchoring, and assess their outputs using hand--depth residuals, object-trajectory smoothness, and control experiments. Across 20 H2O sequences, anchoring reduces the mean of per-sequence median residuals by $23\%$ relative to a single global scale, with improvements on 17 of 20 sequences. The residual is an internal diagnostic because it shares depth with calibration; wrist-excluded, holdout, and perturbation analyses further characterize this dependence (Section~\ref{sec:consistency_eval}).
  \item \textbf{An evaluation with explicit scope and failure analysis.} We report hand accuracy and consistency on 20 H2O sequences (12,032 frames and 21,876 hand samples), camera ATE on three sequences, and an object-trajectory case study on one sequence. Paired camera experiments show that the implemented sensor-depth initialization does not keep depth fixed during sliding-window optimization and yields worse ATE than RGB-only estimation. We distinguish component accuracy, cross-modal diagnostics, and qualitative demonstrations throughout the evaluation (Section~\ref{sec:eval}).
\end{itemize}

\begin{figure}[htbp]
  \centering
  \includegraphics[width=\linewidth]{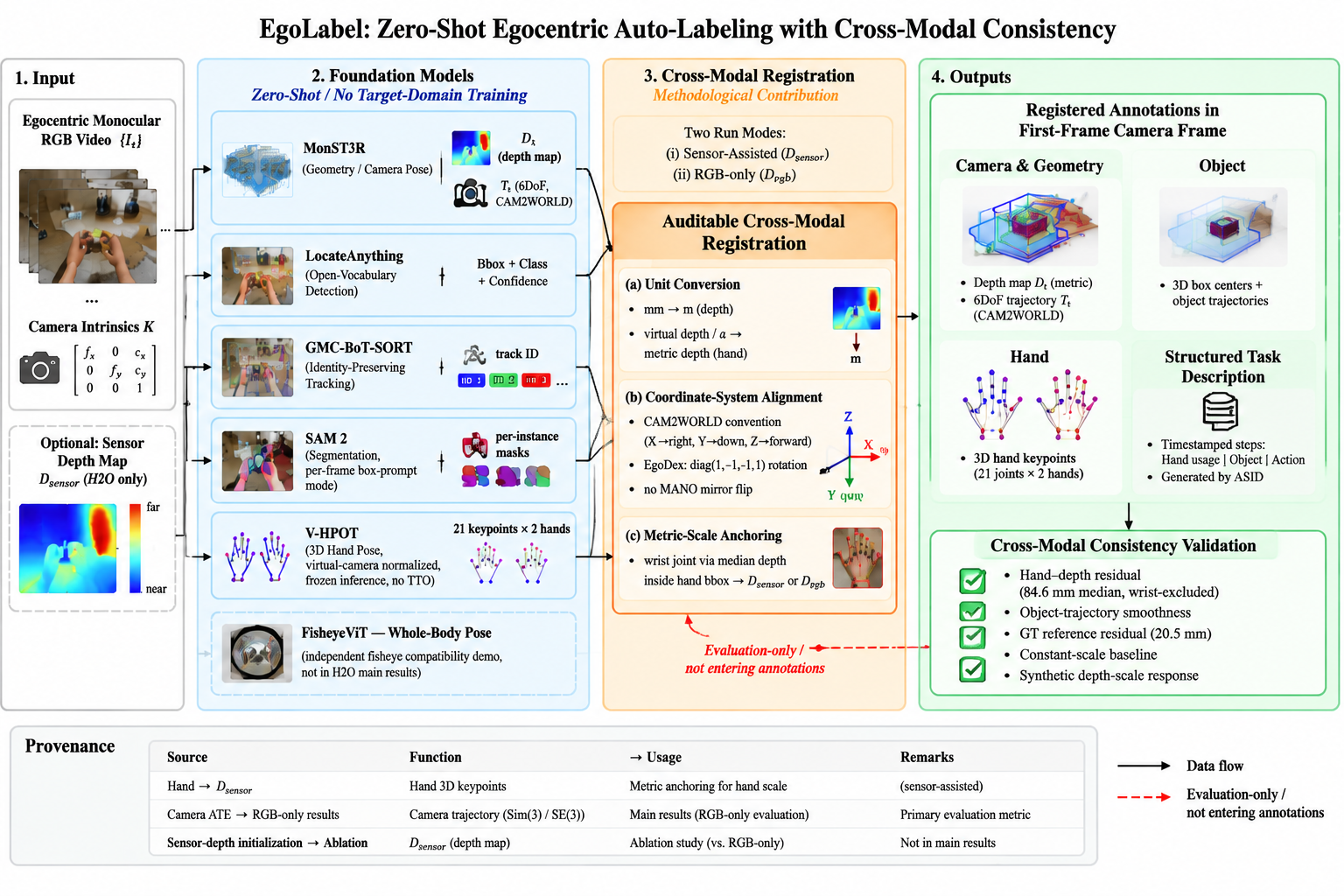}
  \caption{Overall architecture of EgoLabel. Off-the-shelf models produce geometry (MonST3R), object annotations (LocateAnything + GMC-BoT-SORT + SAM~2), and hand poses (V-HPOT). Unit conversion, coordinate alignment, and per-hand depth anchoring connect these streams for joint inspection. Hand--depth residuals, object-trajectory smoothness, and control experiments assess their consistency. FisheyeViT is demonstrated on separate fisheye recordings, and structured manipulation captioning is a downstream application. Metric-scale limitations are detailed in Section~\ref{sec:consistency}.}
  \label{fig:architecture_overview}
\end{figure}

Figure~\ref{fig:architecture_overview} summarizes the pipeline. Section~\ref{sec:related} reviews related work; Section~\ref{sec:methodology} describes the models, registration procedure, and consistency measures; Section~\ref{sec:eval} presents the evaluation, computational cost, and failure analysis; and Section~\ref{sec:conclusion} concludes.

\section{Related Work}
\label{sec:related}

\paragraph{Egocentric perception and datasets.}
Egocentric datasets provide complementary forms of supervision. Ego4D~\cite{grauman2022ego4d} and EPIC-KITCHENS~\cite{damen2022rescaling} offer large-scale activity narration and action labels; AssemblyHands~\cite{kwon2021assemblyhands} and HOT3D~\cite{hot3d2025} provide 3D hand annotations from multi-view or headset capture; and SceneEgo~\cite{sceneego2023} targets whole-body motion. H2O~\cite{kwon2021h2o} provides sensor depth, 6DoF camera poses, hand joints, and object poses in the \emph{same} recording, making it suitable for evaluating agreement across modalities. We use H2O for quantitative evaluation and the in-house ML-EgoDex dataset for compatibility checks (Section~\ref{sec:eval}).

\paragraph{Geometry foundation models.}
DUSt3R~\cite{wang2024dust3r} predicts dense pointmaps for image pairs, providing per-pixel 3D coordinates from which depth, intrinsics, and relative camera poses can be recovered. MonST3R~\cite{zhang2024monst3r} extends this representation to dynamic video through fine-tuning and sliding-window global alignment of poses, pointmaps, and focal length. Its sequence-level alignment provides the geometric backbone of our pipeline. Monocular geometry remains scale-ambiguous, however, so we handle metric anchoring explicitly (Section~\ref{sec:pipeline_integration}). The behavior of sensor-depth initialization in the configuration we use is detailed in Section~\ref{sec:monst3r}.

\paragraph{Hand and whole-body pose estimation.}
MANO-based regressors such as HaMeR~\cite{pavlakos2024hamer} predict parametric hand meshes from image crops. Their crop-normalized camera models require conversion to the recording camera's intrinsics before integration with scene geometry. V-HPOT~\cite{mucha2026towards} instead predicts joint depths in a virtual camera normalized by focal length and image height. It also supports label-free test-time optimization using a 3D-consistency objective over random crops, for which the source paper reports reductions in mean 3D keypoint error of $71\%$ on H2O and $41\%$ on AssemblyHands. We use the frozen checkpoint without test-time optimization and explicitly convert and anchor its virtual-camera outputs (Sections~\ref{sec:vhpot} and~\ref{sec:pipeline_integration}). For whole-body pose, FisheyeViT~\cite{fisheyevit2024} models head-mounted fisheye projection directly with a distortion-aware head. We transform its camera-space keypoints for visualization in a separate compatibility demonstration.

\paragraph{Open-vocabulary detection, tracking, and segmentation.}
Detection has moved from closed-set detectors such as Faster R-CNN~\cite{ren2015faster} to open-vocabulary grounding models such as GroundingDINO~\cite{liu2024groundingdino} and LocateAnything~\cite{locateanything2025}, which take a natural-language prompt and return boxes for categories named only by that prompt. Tracking has moved from SORT~\cite{bewley2016sort} and ByteTrack~\cite{zhang2022bytetrack} to BoT-SORT~\cite{aharon2022botsort}, whose global motion compensation corrects predicted boxes by an affine warp estimated from sparse optical flow before association---a correction that matters because head motion alone can displace the image by hundreds of pixels per frame. Segmentation has moved from per-frame prompting to SAM~2~\cite{ravi2024sam2}, which propagates masks through a video and yields temporally coherent instances (our implementation uses per-frame box prompts with the image predictor rather than video memory propagation). Chained together, these models give identity-preserving, open-vocabulary object masks with no category-specific training. Two properties bear on our scheme: the stream is 2D, so lifting it to 3D requires the depth and pose from the geometry module; and it is identity-preserving across frames, which is what makes the object-trajectory smoothness score of Section~\ref{sec:consistency} meaningful.

\paragraph{Auto-labeling and cross-modal consistency.}
Auto-labeling with foundation models is commonly evaluated per task. Such evaluations establish component accuracy but do not directly test whether independently produced annotations describe a consistent 3D scene. EgoLabel focuses on this cross-modal question by making scale and coordinate conversions explicit and measuring agreement against ground-truth references. To our knowledge, comparable evaluation of integrated egocentric auto-labeling pipelines remains limited. We therefore use internal controls---a ground-truth reference residual, constant-scale anchoring, and synthetic depth-scale perturbations---while acknowledging the absence of a head-to-head system comparison.

\section{Methodology}
\label{sec:methodology}

\subsection{Problem Formulation}
Given an egocentric monocular RGB video $\mathcal{V} = \{I_t\}_{t=1}^{T}$ captured by a head-mounted camera with known intrinsics $\mathbf{K} \in \mathbb{R}^{3\times 3}$---and, optionally, a per-frame sensor depth map---we aim to produce, for every frame $t$, a consistent set of annotations:
\begin{align}
  \mathbf{D}_t &\in \mathbb{R}^{H\times W} &&\text{depth map},\\
  \mathbf{T}_t &\in \mathrm{SE}(3) &&\text{6DoF camera pose (CAM2WORLD)},\\
  \{\mathbf{b}_i, \mathbf{m}_i, c_i\}_{i} &&&\text{object boxes, masks, and labels},\\
  \{\mathbf{J}^{h}_t\}_{h\in\{\mathrm{left},\mathrm{right}\}} \in \mathbb{R}^{2\times 21\times 3} &&&\text{3D hand keypoints (per hand)},\\
  \mathbf{S}_t &&&\text{whole-body pose}.
  \label{eq:outputs}
\end{align}
Depth and hand joints are represented in the per-frame camera coordinate system, and 3D annotations can be transformed to the first-frame camera coordinate system for joint visualization. Sensor depth provides metric scale; RGB-only depth and the hands anchored to it remain scale-ambiguous. Object world trajectories additionally depend on predicted camera translations, whose scale may differ from that of the depth, and are therefore treated as visualizations rather than calibrated world coordinates. $\mathbf{T}_t$ denotes the \emph{world-from-camera} transform (CAM2WORLD), $\mathbf{X}_w = \mathbf{T}_t\,\mathbf{X}_{c_t}$, with $\mathbf{T}_t^{-1}$ mapping world points into the current camera frame. The camera convention is $+X$ right, $+Y$ down, and $+Z$ forward (Section~\ref{sec:monst3r}).

The pipeline uses explicit conversions to reconcile the scales and conventions of independently estimated annotations. In \emph{sensor-assisted} mode, sensor depth initializes MonST3R and is also read directly for hand anchoring and object lifting. In \emph{RGB-only} mode, these operations use MonST3R's predicted depth. Both modes use the same registration procedure and consistency measures; Table~\ref{tab:provenance} identifies the depth and pose sources used by each reported quantity.

\subsection{Foundation Models}
We use off-the-shelf foundation models without target-domain labeled training or fine-tuning. The following sections describe each module and the output conventions relevant to integration.

\subsubsection{Depth Estimation and Camera Pose Estimation}
\label{sec:monst3r}
MonST3R~\cite{zhang2024monst3r} extends DUSt3R's pointmap representation~\cite{wang2024dust3r} to dynamic video (ViT-Large encoder, AsymmetricCroCo3DStereo decoder~\cite{weinzaepfel2022croco}). For a pair of frames it predicts a dense \emph{pointmap} $\mathbf{P}_t \in \mathbb{R}^{H\times W\times 3}$ per view---the 3D coordinates of every pixel, in a common reference frame---plus a confidence map $\mathbf{C}_t$, and is trained with a DUSt3R-style robust regression term,
\begin{equation}
  \mathcal{L}_{\mathrm{pair}} = \sum_{(i,j)} \sum_{t \in \{i,j\}} \big\| \mathbf{C}_t \odot \big( \mathbf{P}_t - \mathbf{X}_t \big) \big\|_{1},
\label{eq:pairwise_loss}
\end{equation}
where $\mathbf{X}_t$ is the target pointmap for view $t$ (the source objective also normalizes the pointmaps and regularizes the confidence map, which we omit here: we do not retrain the model). The two views are related by the relative pose $(\mathbf{R}_{ij}, \mathbf{t}_{ij})$ \emph{through their shared reference frame}, not pixel-by-pixel---the same pixel in two frames is generally not the same scene point, and a dynamic scene is not a single rigid motion. Because monocular pointmaps are scale-ambiguous, the absolute scale is fixed later by the global optimizer or by known depth.

\textbf{Global alignment and known-depth initialization.} A sliding-window global alignment jointly refines poses, per-frame pointmaps and focal length under a temporal-smoothing prior and an optional flow loss. We use resolution $320{\times}176$, window $W{=}30$, overlap $r{=}1/3$, $300$ iterations, flow-loss weight $1.0$ and threshold $30$\,px. Where H2O's 16-bit millimeter depth is available we convert it to meters and write it into the depth parameters,
\begin{equation}
  \texttt{scene.preset\_depthmap}\big(\{\mathbf{D}_t\}, \texttt{requires\_grad=False}\big),
\label{eq:injection_a}
\end{equation}
which in the batchified single-window path also disables the depth gradients. \textbf{In the sliding-window path used for every H2O sequence (more than $30$ frames) it is only a strong initialization}: the window loader re-creates the window's depth \emph{and} focal length as optimizable parameters (\texttt{opt\_im\_depthmaps}, \texttt{opt\_im\_focals}, \texttt{requires\_grad\_(True)} for the first window), re-enabling gradients on the parameters just set, so the injected map is the starting point of the optimization rather than a constraint. Section~\ref{sec:eval} measures the drift ($96\%$ of pixels end more than $1$\,cm away, median relative change $25\%$) and its effect on camera pose. We therefore call this mode \emph{known-depth initialization} and keep the input sensor depth, the window-optimized depth and the RGB-only predicted depth distinct in every table and caption.

MonST3R outputs CAM2WORLD poses in a right-handed camera frame ($+X$ right, $+Y$ down, $+Z$ forward), the frame H2O's stored poses already use, so \emph{no axis flip} is applied for H2O (OpenCV's camera convention is the same). Data stored with the opposite up-axis (the in-house EgoDex-style files) is converted first by \emph{right}-multiplying each pose with $\bar{\mathbf{F}}=\mathrm{diag}(1,-1,-1,1)$, i.e.\ $T' = T\,\bar{\mathbf{F}}$, a $180^{\circ}$ rotation about the $X$ axis that negates the $Y$~and~$Z$ directions in the camera frame while preserving right-handedness; we state the direction because applying it the other way round silently rotates the scene.

\subsubsection{Object Segmentation and Tracking}
\label{sec:object_seg_tracking}

Egocentric manipulation involves dense hand--object interactions where a small set of relevant objects (cups, lids, tools, utensils) appear, disappear, and re-enter the field of view under severe occlusion and camera motion. We therefore require an annotation pipeline that is \emph{open-vocabulary} (not limited to a fixed category set) and \emph{identity-preserving} across frames. Our pipeline chains three models: LocateAnything for detection, BoT-SORT for tracking, and SAM~2 for segmentation.

\paragraph{LocateAnything.}
We adopt LocateAnything~\cite{locateanything2025}, a vision-language grounding model that unifies open-vocabulary detection and fine-grained spatial localization in a single 3B-parameter VLM. Given an input frame $I_t$ resized to a maximum side length of 800\,px, LocateAnything takes a natural-language prompt listing the target categories and returns a set of detections $\mathcal{D}_t = \{(b_i, c_i, s_i)\}_{i=1}^{N_t}$ with bounding boxes, semantic labels, and confidence scores in a single forward pass. We adopt LocateAnything as the primary detector and retain GroundingDINO~\cite{liu2024groundingdino} as a complementary option; both are prompted with the target-category list, and we select LocateAnything for its grounding quality on cluttered scenes (we make no claim that either model was trained for first-person imagery). We apply NMS with an IoU threshold of $0.5$ to suppress duplicate proposals, and discard detections with confidence below $0.35$.

\paragraph{BoT-SORT.}
For identity-preserving tracking we adopt BoT-SORT~\cite{aharon2022botsort}, which augments the two-stage ByteTrack~\cite{zhang2022bytetrack} association with camera-motion compensation (GMC): each track keeps a constant-velocity Kalman state and, before association, its predicted box is corrected by an affine warp estimated from sparse Lucas-Kanade optical flow~\cite{lucas1981iterative} via RANSAC. The correction matters for head-mounted video, where ego motion alone can displace the image by hundreds of pixels per frame and would otherwise fragment identities. Association solves a confidence-modulated IoU assignment with a gate $\tau_{\text{match}}=0.8$, matching high-confidence detections first and then low-confidence ones (those surviving NMS) to recover occluded objects; we disable the appearance ReID head, which is less reliable than spatial overlap in dense, hand-occluded scenes~\cite{wojke2017deepsort}, and terminate tracks after $30$ unmatched frames.

\paragraph{SAM~2.}
With track identities established, we generate per-instance segmentation masks using SAM~2~\cite{ravi2024sam2} (Hiera-L checkpoint, \texttt{sam2.1\_hiera\_large.pt}). Each Kalman-corrected bounding box serves as a box prompt to the image predictor, producing a binary mask $\mathbf{m}_{t,i} \in \{0,1\}^{H\times W}$ for object $i$ at frame $t$. To avoid GPU memory contention between the 3B VLM detector and the 224M-parameter SAM~2 encoder, we explicitly release the detector's VRAM before loading the segmentation model. The final output per frame is an object record $\{track\_id, label, bbox, mask\_path\}$, written incrementally into a structured \texttt{tracks.json} that is consumed by downstream modules (3D lifting and overlay rendering).

\begin{figure}[htbp]
  \centering
  \includegraphics[width=\linewidth]{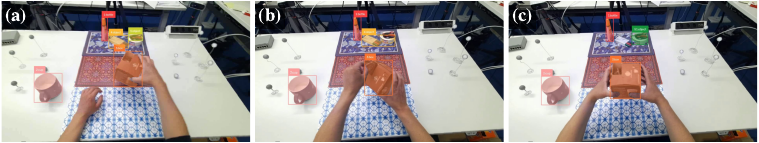}
  \caption{Object tracking and segmentation on H2O h1/0 (frames $0$, $225$, $449$). LocateAnything detects the prompted objects, GMC-BoT-SORT assigns identity-preserving track IDs (colored boxes), and SAM~2 produces per-instance masks overlaid on the frames. The identities are maintained across head motion and partial hand occlusion by the objects.}
  \label{fig:tracking_results}
\end{figure}

\subsubsection{Egocentric 3D Hand Pose Estimation}
\label{sec:vhpot}
For 3D hand pose estimation we adopt \textbf{V-HPOT}~\cite{mucha2026towards}, which is camera-agnostic by construction: it predicts keypoint depths in a \emph{virtual camera} normalized by focal length and image height, so it does not overfit the recording camera's intrinsics. With focal length $f$ (we use $f_x$), image height $H$ (the network input height) and virtual parameters $(f_v,H_v)$,
\begin{equation}
  \alpha = \frac{f_v}{f}\cdot\frac{H}{H_v},\qquad
  z_{\text{virtual}} = \alpha\, z_{\text{metric}},\qquad
  z_{\text{metric}} = z_{\text{virtual}}/\alpha,
\label{eq:virtual_cam}
\end{equation}
i.e.\ the model's depth is the metric depth scaled by $\alpha$. We follow the source paper's direction and record in the supplementary material which direction our released checkpoint uses: applying the factor the other way round would only rescale the whole skeleton by a constant, so scale-invariant metrics would not reveal the error. The formula uses $f$ at the resolution where the keypoint pixel coordinates are expressed; if the input image is resized to $H$ the focal length must be scaled accordingly. Keypoints are unprojected with the dataset intrinsics,
\begin{equation}
  \mathbf{X} = \mathbf{K}^{-1}\,\begin{bmatrix}\mathbf{x}\\1\end{bmatrix}\cdot z,
\label{eq:unproject}
\end{equation}
which gives $X=(u-c_x)z/f_x$ and $Y=(v-c_y)z/f_y$, with $+Y$ \emph{down} in the image, consistent with the camera frame defined above. The backbone is an EfficientNet-V2-S with a depth head and a hand head that predicts $21$ heatmaps and per-joint depths per hand plus a handedness score; on two-hand data it returns $42$ keypoints in each frame's camera frame. The model also offers a self-supervised test-time optimization (TTO) that refines it without labels; our released configuration does \emph{not} use TTO---the checkpoint runs frozen, one forward pass per batch---so every hand number here is without TTO and we make no claim about its reported gains (Section~\ref{sec:related}).

\paragraph{Integration.}
For each frame V-HPOT consumes the RGB image resized to $224{\times}224$ and returns $42$ keypoints in the virtual camera space, which require conversion and scale correction before combination with depth. We register them in two steps: divide the virtual depth by $\alpha$ (Eq.~\eqref{eq:virtual_cam}) using the dataset intrinsics, then correct the residual absolute scale by anchoring the wrist to the depth (Section~\ref{sec:pipeline_integration}). Because V-HPOT outputs camera-space coordinates (unlike MANO-space predictors), no left/right X-mirroring is needed. Over the 20 H2O sequences it attains a PA-MPJPE of $14.53\,{\pm}\,1.29\,\mathrm{mm}$ (21,876 hand samples); we do not directly compare this result with HaMeR because the quoted $18.27\,\mathrm{mm}$ comes from a different protocol.

\subsubsection{Whole-Body Pose Estimation}
\label{sec:whole_body_estimation}

\begin{figure}[htbp]
  \centering
  \includegraphics[width=\linewidth]{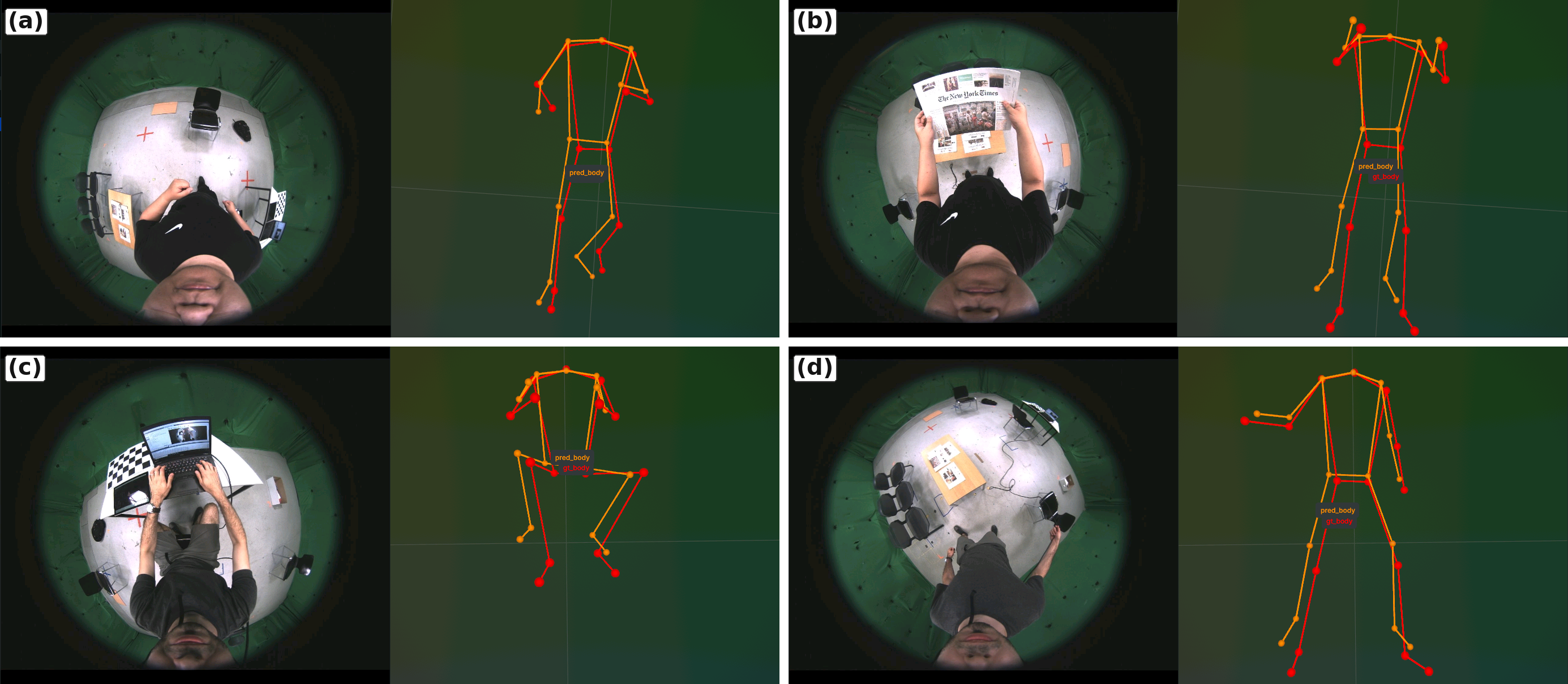}
  \caption{Qualitative whole-body results from the FisheyeViT stream. Each panel pairs a head-mounted fisheye frame (left) with the reconstructed 3D skeleton in the shared metric frame (right); predicted skeletons are orange and, where available, ground-truth skeletons are red. (a)~Prediction only; (b)--(d)~prediction overlaid with ground truth. The skeleton is mapped into the shared frame by the camera-to-world transform (Section~\ref{sec:whole_body_estimation}); this is a compatibility demonstration on separate recordings, not a joint H2O result.}
  \label{fig:whole_body}
\end{figure}

We include the wearer's body $\mathbf{S}_t$ (Eq.~\eqref{eq:outputs}) for completeness, using \textbf{FisheyeViT}~\cite{fisheyevit2024}, a ViT-based estimator for head-mounted fisheye cameras that models the fisheye projection explicitly instead of undistorting the input; we run its $256{\times}256$ checkpoint (EgoBody/EgoWholeBody). H2O uses a pinhole RealSense camera, so this is an \emph{independent compatibility demonstration} on SceneEgo's fisheye sequences, not a component of the H2O results, and we do not re-evaluate it (the source paper reports $\text{MPJPE}_{\text{body}}=72.3\,\mathrm{mm}$ and $\text{PA-MPJPE}_{\text{body}}=48.7\,\mathrm{mm}$ on SceneEgo\_test). Its keypoints are mapped into the shared frame by the camera-to-world transform, $\mathbf{J}^w_j=\mathbf{T}_{\text{cam}}\,\mathbf{J}^{\text{cam}}_j$; because the recording differs, this stream is not fused with the H2O streams and we claim only that the same registration procedure can place a body skeleton in the shared frame. Figure~\ref{fig:whole_body} shows representative outputs.

\subsection{Pipeline Integration and Engineering Considerations}
\label{sec:pipeline_integration}
Integrating the models requires explicit handling of units, camera conventions, and scale. This section describes the data flow and the transformations used to connect their outputs.

\paragraph{End-to-end data flow.}
The integrated pipeline (implemented by \texttt{h2o\_adapter.py}) proceeds as follows:
\begin{enumerate}
  \item \textbf{Frames and intrinsics.} Raw H2O RGB frames ($1280{\times}720$) are normalized to the canonical naming \texttt{\{index:05d\}.jpg}; the camera intrinsics $\mathbf{K}$ are read from the dataset, and every downstream resolution change is an explicit resize (the MonST3R input is $320{\times}176$ and the V-HPOT input $224{\times}224$, neither of which is a proportional scaling of $1280{\times}720$, so 2D quantities are mapped through the corresponding scale factors and depths are sampled nearest-neighbor on the resized grid).
  \item \textbf{Depth and camera pose (MonST3R).} MonST3R runs at $320{\times}176$ with sliding-window global alignment; the known H2O depth initializes the depth parameters (Eq.~\eqref{eq:injection_a}), producing the depth map $\mathbf{D}_t$ and poses $\mathbf{T}_t$ stored in \texttt{pred\_poses.hdf5}.
  \item \textbf{Object detection, tracking, and segmentation.} LocateAnything performs open-vocabulary detection, GMC-BoT-SORT associates detections into identity-preserving tracks, and SAM~2 produces per-instance masks (Section~\ref{sec:object_seg_tracking}).
  \item \textbf{3D hand pose (V-HPOT).} Each frame is resized to $224{\times}224$ and passed to V-HPOT, producing $42$ keypoints (21 per hand) with handedness scores, stored as \texttt{hand\_tracks.json}.
  \item \textbf{Whole-body pose (FisheyeViT).} The wearer's body is estimated separately and mapped into the same frame via $\mathbf{J}^w = \mathbf{T}_{\text{cam}}\,\mathbf{J}^{\text{cam}}$ (Section~\ref{sec:whole_body_estimation}).
  \item \textbf{Overlay and 3D visualization.} An annotated overlay video is rendered and all entities are projected into the shared world frame to produce the final RRD visualization.
\end{enumerate}

\begin{figure}[htbp]
  \centering
  \includegraphics[width=\linewidth]{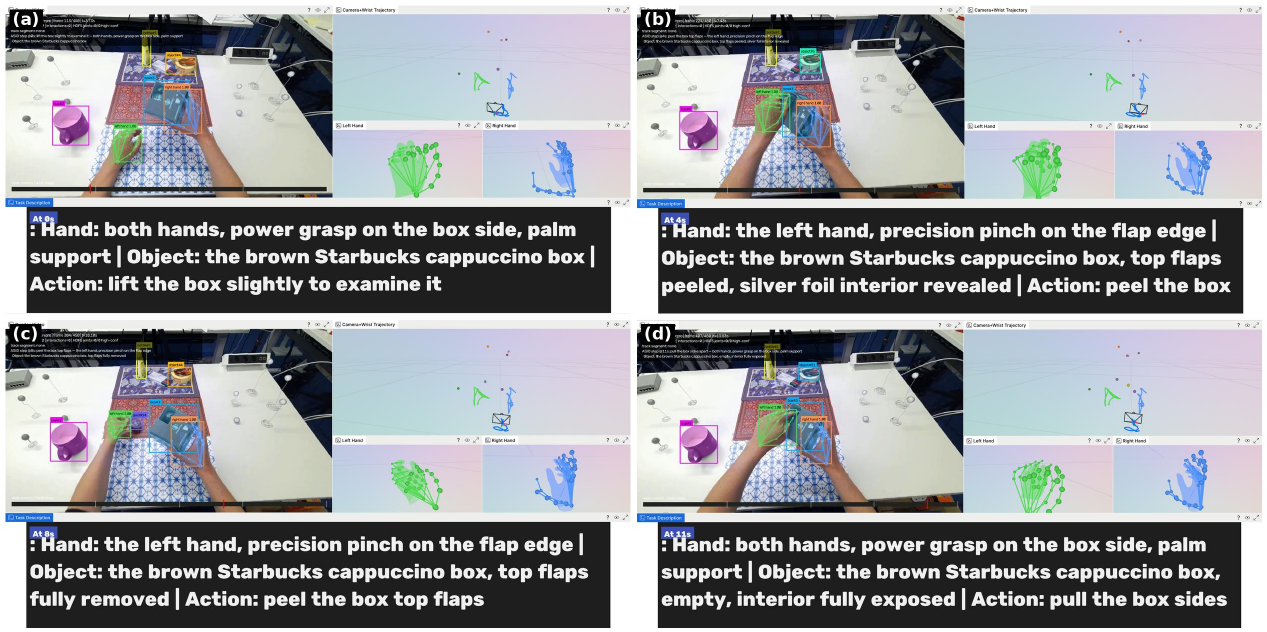}
  \caption{Final end-to-end output of EgoLabel on H2O \texttt{h1/0}, as rendered in the Rerun viewer at four stages of the manipulation: (a)~lifting the box to examine it, (b)~peeling the top flaps, (c)~removing the flaps, and (d)~pulling the box sides. Each panel is one screenshot of the integrated viewer and simultaneously shows the object stream (LocateAnything detections, GMC-BoT-SORT track IDs, and SAM~2 masks, drawn as colored boxes on the overlay video), the 3D hand prediction (left/right V-HPOT skeletons and the camera+wrist trajectory in 3D), and the ASID task caption (bottom panel: hand usage, manipulated object, and action). Object, hand, and camera annotations are produced by independent zero-shot models; the ASID caption (bottom panel) is a GT-assisted downstream consumer (Section~\ref{sec:captioning}).}
  \label{fig:pipeline_output}
\end{figure}

\paragraph{Unit conversion.}
The modules output different units, reconciled before fusion. H2O depth is 16-bit millimeters, divided by $1000$ to meters before the known-depth initialization; V-HPOT keypoints are converted with Eq.~\eqref{eq:virtual_cam} using the dataset intrinsics before scale correction; and the evaluation-only trajectory alignment is a 7-DoF similarity solved in closed form (Umeyama): with covariance $\mathbf{H}=\sum_t(\mathbf{q}_t-\boldsymbol{\mu}_q)(\mathbf{p}_t-\boldsymbol{\mu}_p)^{\!\top}=\mathbf{U}\boldsymbol{\Sigma}\mathbf{V}^{\!\top}$,
\begin{equation}
  \mathbf{R} = \mathbf{U}\,\mathrm{diag}\big(1,1,\det(\mathbf{U}\mathbf{V}^{\!\top})\big)\,\mathbf{V}^{\!\top}, \quad
  s = \frac{\mathrm{tr}\big(\boldsymbol{\Sigma}\,\mathrm{diag}(1,1,\det(\mathbf{U}\mathbf{V}^{\!\top}))\big)}{\sum_t \|\mathbf{p}_t-\boldsymbol{\mu}_p\|^2}, \quad
  \mathbf{t} = \boldsymbol{\mu}_q - s\,\mathbf{R}\,\boldsymbol{\mu}_p,
  \label{eq:umeyama}
\end{equation}
the $\mathrm{Sim}(3)$ solution minimizing $\sum_t \| s\,\mathbf{R}\,\mathbf{p}_t + \mathbf{t} - \mathbf{q}_t \|^2$; it is applied \emph{only} to report ATE. \textbf{Registration uses no ground-truth poses}: all streams are normalized to the first-frame camera space with $\mathbf{R}=\mathbf{I}$ ($\mathbf{T}_t \leftarrow \mathbf{T}_0^{-1}\mathbf{T}_t$; the reference-referenced alignment of Eq.~\eqref{eq:umeyama} is evaluation-only). In the sensor-assisted mode, hand anchoring (Eq.~\eqref{eq:scale_correction}) and object lifting read $\mathbf{D}_{\text{sensor}}$; the known-depth initialization (Section~\ref{sec:monst3r}) is part of this mode. The RGB-only variant uses $\mathbf{D}_{\text{rgb}}$.

\paragraph{Coordinate-system conventions.}
MonST3R and H2O share one right-handed frame ($+X$ right, $+Y$ \emph{down}, $+Z$ forward, CAM2WORLD; also OpenCV's camera convention), so \emph{no flip} is applied for H2O; the in-house EgoDex-style files store the opposite up-axis and are converted first by $\bar{\mathbf{F}} = \mathrm{diag}(1,-1,-1,1)$, i.e.\ $T' = T\,\bar{\mathbf{F}}$ (a $180^{\circ}$ rotation about $X$). MANO-based regressors output the left hand mirrored, which V-HPOT avoids by working directly in camera space.

\paragraph{Metric-scale correction of the hand skeleton.}
V-HPOT keypoints keep a residual absolute-scale ambiguity, which we remove by anchoring the wrist to the depth map---sampling the median depth over the hand box and scaling the skeleton,
\begin{equation}
  s_{\mathrm{depth}} = \frac{\mathrm{median}\big\{ \mathbf{D}_t(u,v) : (u,v) \in \mathrm{bbox}^{\mathrm{hand}}_t \big\}}{z^{\mathrm{wrist}}_t}, \qquad
  \tilde{\mathbf{J}}_t = s_{\mathrm{depth}}\, \mathbf{J}_t.
  \label{eq:scale_correction}
\end{equation}
The depth ratio can differ substantially from one, so we use a wide clamp, $[10^{-6},10^{6}]$, to avoid unintended truncation. The pipeline applies Gaussian temporal smoothing with $\sigma=2$ frames; the resulting scale factor therefore approximates the per-frame ratio. Under occlusion, the box median may sample the background or a manipulated object rather than the hand, motivating a hand-mask-based alternative (Section~\ref{sec:failure}).

\paragraph{Operational details.}
The known-depth initialization requires meters: the values go into log-depth parameters, so a millimeter map is silently mis-scaled. Depth correction is unnecessary for the scale-invariant hand metrics but required for fusion, so that hands, objects and trajectory share one frame; all entities are logged under distinct paths (\texttt{world/hands}, \texttt{world/gt\_hand}, \texttt{world/objects}, \texttt{world/trajectory\_*}) for frame-by-frame inspection.

\subsection{Cross-Modal Consistency Measures}
\label{sec:consistency}
We assess cross-modal agreement using a hand--depth residual and an object-trajectory smoothness score. The residual reuses the depth map that calibrates hand scale, so it measures internal consistency rather than independently validating the anchoring. Most hand and object results are sensor-assisted; RGB-only results are labeled separately. Table~\ref{tab:provenance} specifies the depth source, pose source, and coordinate frame for each quantity, distinguishing input sensor depth from depth refined by MonST3R.

\begin{table}[htbp]
  \centering
  \caption{Provenance of the geometry behind each reported quantity. ``Sensor depth'' is the dataset's depth map converted to meters and resampled to $320{\times}176$ (nearest neighbor); it never passes through MonST3R. ``Window-optimized'' is the depth after the sliding-window alignment of Section~\ref{sec:monst3r}.}
  \label{tab:provenance}
  \small
  \begin{tabular}{@{}l l l l@{}}
    \toprule
    Quantity & Depth source & Pose source & Frame \\
    \midrule
    Hand scale anchoring, Eq.~\eqref{eq:scale_correction} & sensor depth, box median & --- & camera \\
    Hand--depth residual, Eq.~\eqref{eq:hand_depth_resid} & the same map, nearest sample & --- & camera \\
    Object-trajectory smoothness, Eq.~\eqref{eq:obj_smooth} & sensor depth, box median & predicted $\mathbf{T}_t$ & world \\
    Object single-object case study & sensor depth, box center & --- & camera \\
    Camera ATE, main (RGB-only) & MonST3R predicted & MonST3R predicted & first-frame camera \\
    Camera ATE, ablation & window-optimized (initialized) & MonST3R predicted & first-frame camera \\
    \bottomrule
  \end{tabular}
\end{table}

The object smoothness score is computed in world coordinates using predicted poses, whereas the single-object trajectory case study is evaluated in camera coordinates without predicted poses. Their scale assumptions differ, so we report them separately.

We distinguish four depth maps: $D_{\text{sensor}}$, the sensor measurement converted to meters; $D_{\text{init}}$, the same map written into MonST3R; $D_{\text{opt}}$, the depth after window optimization; and $D_{\text{rgb}}$, the depth predicted from RGB alone. Optimization changes $D_{\text{init}}$, so it must not be conflated with $D_{\text{opt}}$ (Section~\ref{sec:eval}). The 20-sequence hand study uses $D_{\text{sensor}}$ directly for both anchoring and evaluation, independently of MonST3R. World-frame object smoothness combines sensor-depth points with predicted camera translations that may have a different scale; we consequently interpret it as a relative stability diagnostic.

\paragraph{Hand--depth residual.}
After metric-scale correction (Eq.~\eqref{eq:scale_correction}), each V-HPOT joint $\tilde{\mathbf{J}}_j$ lies in metric camera space. We reproject it to the image plane using the intrinsics $\mathbf{K}$ and compare its depth with the depth map $\mathbf{D}_t$ ($\mathbf{D}_{\text{sensor}}$ in the 20-sequence study, $\mathbf{D}_{\text{rgb}}$ in the RGB-only variant):
\begin{equation}
  r_j = \Big| \tilde{z}_j - \mathbf{D}_t\!\left(\pi(\tilde{\mathbf{J}}_j)\right) \Big|,
  \label{eq:hand_depth_resid}
\end{equation}
where $\pi(\cdot)$ denotes perspective projection and $\mathbf{D}_t$ is sampled with nearest-neighbor interpolation at the MonST3R resolution. A small $r_j$ indicates agreement between the estimated joint depth and the sampled surface depth at its projected location.

\paragraph{Object-trajectory smoothness.}
For each tracked object we back-project its 2D box center using $\mathbf{D}_t$ and the per-frame pose $\mathbf{T}_t$ to obtain a 3D box-center proxy $\mathbf{c}_t$, and per frame compute the second difference,
\begin{equation}
  a_t = \frac{\big\| \mathbf{c}_{t+\Delta t} - 2\mathbf{c}_t + \mathbf{c}_{t-\Delta t} \big\|}{\Delta t^{2}}, \quad \Delta t = 1\ \text{frame},
  \label{eq:obj_smooth}
\end{equation}
aggregated as the median over valid object-frame samples. Because egocentric camera motion would otherwise inject large spurious accelerations, this score provides a temporal-stability diagnostic affected by tracking, depth, and camera estimation; a large value on a static object can indicate a registration error, though a constant-bias trajectory can also produce a low score. At $f$ frames per second the per-frame second difference corresponds to an acceleration of $f^{2}a_t$ (for H2O's $30$\,fps, $900\,a_t$), so we report both the raw per-frame value and $\mathrm{mm/s^2}$ where relevant.

\paragraph{Hand--object proximity (defined, not evaluated).}
A minimum distance between fingertip joints and the back-projected object mask would test whether the hand and object streams agree during a grasp; we define it for completeness but do not evaluate it, so it is \emph{not} part of the validated protocol of this paper.

\paragraph{Naive stacking vs.\ consistency scheme.}
\emph{Naive stacking} combines model outputs in their native units, leaving V-HPOT in virtual-camera units and depth in meters. This unit mismatch makes the resulting distances uninterpretable. We report it only to illustrate the need for conversion; the meaningful control for per-hand anchoring is a single global scale, evaluated in Section~\ref{sec:consistency_eval}.

\begin{figure}[htbp]
  \centering
  \includegraphics[width=0.95\linewidth]{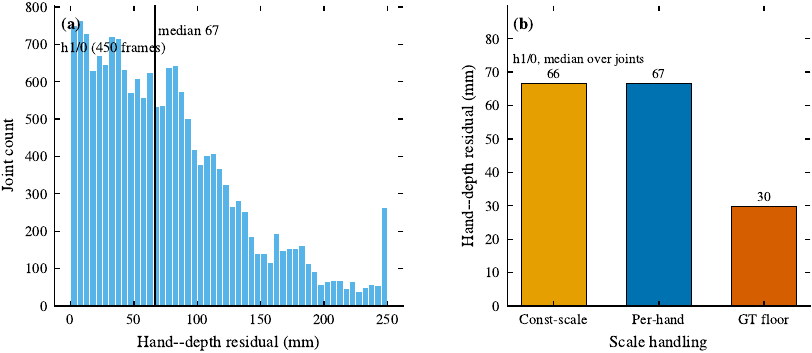}
  \caption{Cross-modal consistency of the V-HPOT hand stream against the depth map on H2O \texttt{h1/0}. (a) Distribution of the per-joint hand--depth residual; the solid line is the median. (b) Median residual for a single global scale, our per-hand anchoring, and the reprojected ground-truth hands. The figure shows a representative sequence; across the 20 sequences the per-sequence median ranges from $64.2$ to $119.7\,\mathrm{mm}$, and the pooled medians are in Table~\ref{tab:consistency}. The residual is a consistency check (does the fused skeleton lie on the reconstructed surface?), not an independent accuracy metric.}
  \label{fig:consistency}
\end{figure}

\subsection{Downstream Application: Structured Manipulation Captioning}
\label{sec:captioning}
We illustrate downstream use through structured manipulation captioning for robot policy learning. ASID-Captioner, a video MLLM based on Qwen2.5-Omni-3B~\cite{qwen2025omni} and trained on attribute-structured audiovisual instructions, receives the assembled RGB video and produces timestamped manipulation steps. To improve left/right attribution, we use hand motion to distinguish a \emph{holder} hand that stabilizes the object from a \emph{mover} hand that performs the action. On H2O, this attribution uses ground-truth hand poses, making the demonstration \emph{GT-assisted}. Predicted timestamps are linearly rescaled to the video duration, and references to audio and hallucinated content are removed. Figure~\ref{fig:pipeline_output}(a)--(d) shows the resulting captions alongside the geometry annotations. Captioning is a qualitative downstream application and is not quantitatively evaluated.

\paragraph{Prompt design: from scene description to manipulation steps.}
Generic video captions tend to emphasize scene appearance rather than individual manipulation steps. Our robot-oriented prompt specifies three fields: \emph{Hand} (left/right identity, grasp type, and contact points), \emph{Object} (identity, appearance, location, and state), and \emph{Action} (e.g., reach, grasp, lift, rotate, peel, open, insert, or place). It requests separate descriptions of the two hands and a new step when the object state changes, while excluding background, camera, atmosphere, and audio descriptions. This format organizes captions around manipulation events for use alongside the 3D annotations.

\section{Evaluation}
\label{sec:eval}
\subsection{Dataset Selection}
We use H2O for quantitative evaluation and additional datasets for qualitative compatibility checks. The evaluation separates hand accuracy and consistency, camera trajectories, and object-stream case studies.

\paragraph{H2O (Hands and Objects in Egocentric action).}
H2O~\cite{kwon2021h2o} records two-hand manipulation with a head-mounted RealSense camera at $30$ fps and $1280{\times}720$ resolution. It provides synchronized RGB and 16-bit depth images (in millimeters), camera intrinsics and 6DoF poses, 21 keypoints per hand, MANO~\cite{romero2017embodied} parameters, and object poses. We evaluate hand accuracy and hand--depth consistency on 20 \texttt{subject1\_ego} sequences: \texttt{h1/0}--\texttt{h1/7}, \texttt{h2/0}--\texttt{h2/7}, and \texttt{o1/0}--\texttt{o1/3}, totaling $12{,}032$ frames and $21{,}876$ hand samples. These runs use sensor depth directly and evaluate hand registration independently of camera estimation and object lifting. Camera ATE is evaluated on three sequences (\texttt{h1/0}, \texttt{k1/0}, and \texttt{o1/0}; $1{,}233$ frames), while object smoothness and the single-object case study use \texttt{h1/0}, the only sequence with full-pipeline object tracks. Table~\ref{tab:coverage} summarizes the coverage. We report camera results with and without sensor-depth initialization, taking RGB-only as the main camera configuration, and provide an RGB-only hand--depth diagnostic on two sequences.

\paragraph{Compatibility datasets.}
ML-EgoDex (in-house; \texttt{.mp4} plus \texttt{.hdf5} intrinsics and hand/object keypoints; no sensor depth) and HOT3D~\cite{hot3d2025} (Project Aria, 3D hand annotations and poses; the source of V-HPOT's pre-training) are used only to check \emph{compatibility} of the pipeline without sensor depth and of the hand module across cameras. We report no quantitative numbers on either.

Table~\ref{tab:dataset_coverage} summarizes the available annotations and the role of each dataset. H2O supplies the joint depth, camera, hand, and object annotations needed for our quantitative study; ML-EgoDex and HOT3D support compatibility checks, and SceneEgo supplies the separate whole-body demonstration.

\begin{table}[htbp]
  \centering
  \caption{Annotation coverage and role of the datasets. $\checkmark$ = ground truth available; --- = not available. ``Quant.'' indicates whether the dataset yields quantitative results in this work.}
  \label{tab:dataset_coverage}
  \begin{tabular}{lcccccc}
    \toprule
    Dataset & Depth & Cam.\ pose & Hand & Object & Body & Quant. \\
    \midrule
    H2O~\cite{kwon2021h2o} & $\checkmark$ & $\checkmark$ & $\checkmark$ & $\checkmark$ & --- & $\checkmark$ \\
    ML-EgoDex (in-house) & --- & --- & $\checkmark$ & $\checkmark$ & --- & --- \\
    HOT3D~\cite{hot3d2025} & --- & $\checkmark$ & $\checkmark$ & $\checkmark^{\dag}$ & --- & --- \\
    SceneEgo~\cite{sceneego2023} & --- & --- & --- & --- & $\checkmark$ & --- \\
    \bottomrule
  \end{tabular}
  \par\smallskip
  {\footnotesize $^{\dag}$HOT3D provides object pose and mesh annotations~\cite{hot3d2025}; our compatibility check does not use them.}
\end{table}

\begin{table}[htbp]
  \centering
  \caption{Coverage of the quantitative results in Section~\ref{sec:eval}. Hand accuracy and hand--depth consistency are measured on 20 sequences; camera ATE on three; object-trajectory smoothness requires the full tracking run and is validated on \texttt{h1/0}. RGB-only variants are listed separately.}
  \label{tab:coverage}
  \small
  \begin{tabular}{@{}l l l@{}}
    \toprule
    Result & Sequences & Frames / samples \\
    \midrule
    Hand pose (PA-MPJPE, scale-aligned) & 20 & $12{,}032$ frames / $21{,}876$ samples \\
    Hand--depth residual & 20 & $12{,}032$ frames / $429{,}896$ joints \\
    Camera ATE ($\mathrm{Sim}(3)$, $\mathrm{SE}(3)$) & 3 (\texttt{h1/0}, \texttt{k1/0}, \texttt{o1/0}) & $1{,}233$ frames \\
    Camera ATE, no injection (RGB-only) & 3 (\texttt{h1/0}, \texttt{k1/0}, \texttt{o1/0}) & $1{,}233$ frames \\
    Hand--depth residual, no injection & 2 (\texttt{h1/0}, \texttt{o1/0}) & --- \\
    Object-trajectory smoothness / 1-object case study & \texttt{h1/0} & $450$ frames / $1{,}925$ detections \\
    Object detection/tracking/segmentation & \texttt{h1/0} (stability) & $450$ frames / $9$ tracks \\
    Whole-body & fisheye sequences (qual.) & --- \\
    Captioning & \texttt{h1/0} (qual.) & --- \\
    \bottomrule
  \end{tabular}
\end{table}

\subsection{Component Evaluation and Sanity Checks}
\label{sec:reproduction_check}
We first evaluate the pretrained components on the selected H2O subsets to establish a baseline for the cross-modal analysis. These results characterize the models within our pipeline; differences in datasets and protocols prevent direct claims of reproducing their published benchmarks. Table~\ref{tab:module_eval} summarizes camera and hand accuracy. Object-stream stability and whole-body compatibility are discussed separately.

\begin{table}[htbp]
  \centering
  \caption{Per-module accuracy measured with the pipeline on H2O with off-the-shelf models, reported as a component sanity check rather than a new benchmark claim. Camera ATE is averaged over three sequences (\texttt{h1/0}, \texttt{k1/0}, \texttt{o1/0}; 1,233~frames); hand metrics over 20 sequences (h1/0--h1/7, h2/0--h2/7, o1/0--o1/3; 12,032~frames, 21,876 hand samples). Hand metrics are means over hand samples within a sequence, then averaged across sequences; $\pm$ denotes sample standard deviation across sequences, not across samples. V-HPOT is run frozen (no test-time optimization); the source paper's $71\%$ figure is for its TTO variant and is not reproduced here.}
  \label{tab:module_eval}
  \small
  \begin{tabular}{@{}l l l c@{}}
    \toprule
    Module & Metric & Reported in paper & EgoLabel (H2O) \\
    \midrule
    MonST3R & Camera ATE, $\mathrm{Sim}(3)$ & Competitive vs.\ DUSt3R/CasualSAM & $1.40$\,${\pm}$\,$0.29\,\mathrm{cm}$$^{\S}$ \\
    MonST3R & Camera ATE, $\mathrm{SE}(3)$ & --- & $1.60$\,${\pm}$\,$0.51\,\mathrm{cm}$$^{\S}$ \\
    V-HPOT (frozen) & PA-MPJPE & cross-domain gains with TTO & $14.53$\,${\pm}$\,$1.29\,\mathrm{mm}$ \\
    V-HPOT (frozen) & Scale-aligned MPJPE & --- & $26.46$\,${\pm}$\,$4.40\,\mathrm{mm}$ \\
    V-HPOT (frozen) & Absolute (metric) MPJPE & --- & $119.1$\,${\pm}$\,$31.3\,\mathrm{mm}$$^{\ddag}$ \\
    \bottomrule
  \end{tabular}

  \vspace{2pt}
  {\footnotesize $^{\ddag}$Absolute MPJPE compares the anchored (metric) hands with GT without any Procrustes alignment; see Section~\ref{sec:eval}. $^{\S}$RGB-only (no injected depth). The sensor-depth-injection ablation is worse ($2.11\pm1.03$ / $3.81\pm0.62\,\mathrm{cm}$) in a configuration where depth is not kept fixed during window-wise alignment (Section~\ref{sec:eval}). Separately, a MANO-based regressor (HaMeR~\cite{pavlakos2024hamer}) is reported at $18.27\,\mathrm{mm}$ PA-MPJPE on \texttt{h1/0} under a different protocol; we do not treat it as a controlled comparison.}
\end{table}

\begin{figure}[htbp]
  \centering
  \includegraphics[width=0.9\linewidth]{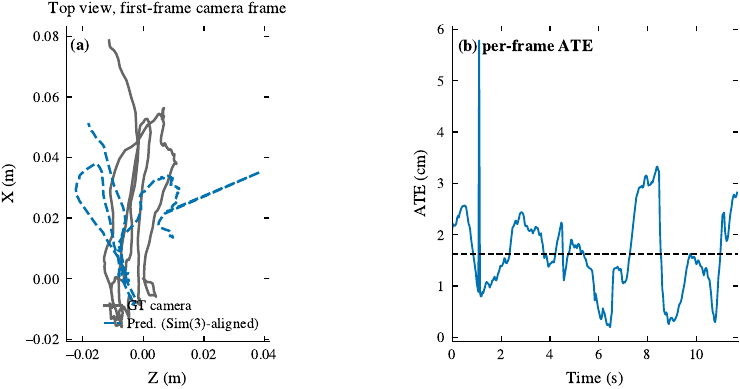}
  \caption{Camera trajectory on H2O \texttt{k1/0} for the \emph{sensor-depth-injection ablation} (the RGB-only trajectory, which we take as the main camera result, is summarized in Table~\ref{tab:ate_noinj}), in the first-frame camera frame. (a) Top view: ground-truth (gray) and MonST3R trajectory after 7-DoF $\mathrm{Sim}(3)$ alignment (blue). (b) Per-frame ATE (blue) with the mean as a dashed line. The injected run recovers the trajectory \emph{shape} well, with the residual global-scale error left by the injection acting only as an initialization rather than a constraint (recomputed with the released estimator: $\mathrm{Sim}(3)$ ATE $1.76\,\mathrm{cm}$, fitted scale $0.45$, strict $\mathrm{SE}(3)$ $3.12\,\mathrm{cm}$).}
  \label{fig:ate}
\end{figure}

\paragraph{Camera pose.}
In the MonST3R paper~\cite{zhang2024monst3r}, camera pose is evaluated on TUM-dynamics and ScanNet. Within EgoLabel, MonST3R runs in the two modes of Section~\ref{sec:methodology} at $320{\times}176$, window $30$, overlap $1/3$, $300$ iterations; Table~\ref{tab:ate_noinj} reports both on the three H2O ATE sequences (\texttt{h1/0}, \texttt{k1/0}, \texttt{o1/0}; $1{,}233$ frames), paired on the same frames and computed with a common evaluation implementation using similarity or rigid alignment of camera translations. \textbf{We take the RGB-only mode as the main camera result}: $1.40\,{\pm}\,0.29\,\mathrm{cm}$ ($\mathrm{Sim}(3)$) and $1.60\,{\pm}\,0.51\,\mathrm{cm}$ ($\mathrm{SE}(3)$), with fitted scales $0.85/1.34/1.78$. Since H2O's poses are metric, the $\mathrm{SE}(3)$ value is the stricter measure, and the small gap indicates a limited effect of scale alignment on ATE for these sequences; the fitted scales still vary across sequences, so RGB-only estimation yields lower trajectory error than the initialization ablation without establishing reliable metric geometry. For reference, the unaligned trajectory differs from GT by $21.54\,\mathrm{cm}$ on \texttt{h1/0}, an unaligned discrepancy rather than an aligned ATE; Figure~\ref{fig:ate} shows the injected trajectory and its per-frame error for \texttt{k1/0}.

\paragraph{Depth changes during sliding-window optimization.}
The sensor-assisted mode writes sensor depth with \texttt{requires\_grad=False} (Eq.~\eqref{eq:injection_a}), but the sliding-window loader re-creates depth and focal length as optimizable parameters. The sensor map therefore serves as an initialization: on \texttt{h1/0}, $96\%$ of pixels differ from the input by more than $1$\,cm after optimization, with a median relative change of $25\%$. The fitted geometry scales range from $0.17$ to $0.45$, and camera ATE is worse than RGB-only on all three sequences ($2.11\,{\pm}\,1.03$ / $3.81\,{\pm}\,0.62\,\mathrm{cm}$ for $\mathrm{Sim}(3)$ / $\mathrm{SE}(3)$). Initializing focal length from dataset intrinsics does not resolve the discrepancy. We retain this configuration as an ablation. Hand--depth residuals and the single-object case study use sensor depth directly in camera coordinates and are unaffected by this optimization; world-frame object smoothness depends on the predicted poses and inherits their scale limitations (Table~\ref{tab:provenance}).

\begin{table}[htbp]
  \centering
  \caption{Camera ATE (cm) on the three H2O ATE sequences, paired on the same sequences and frame sets and computed with the same evaluation script (\texttt{scripts/eval\_ate.py}), using similarity or rigid alignment of camera translations. ``Inj.'' is the sensor-depth initialization (the released sensor-assisted mode) with the focal either optimized (free) or initialized from the dataset intrinsics (focal-init, not pinned); ``RGB-only'' uses MonST3R's predicted depth and is the main camera result. Fitted $\mathrm{Sim}(3)$ scale in parentheses.}
  \label{tab:ate_noinj}
  \small
  \begin{tabular}{@{}lcccccc@{}}
    \toprule
    Seq. & RGB $\mathrm{Sim}(3)$ (scale) & RGB $\mathrm{SE}(3)$ & inj.\ free $\mathrm{Sim}(3)$ (scale) & inj.\ $\mathrm{SE}(3)$ & inj.\ focal-init $\mathrm{Sim}(3)$ \\
    \midrule
    \texttt{h1/0} & $1.08$ ($0.85$) & $1.09$ & $1.29$ ($0.17$) & $4.00$ & $1.34$ \\
    \texttt{k1/0} & $1.51$ ($1.34$) & $1.62$ & $1.76$ ($0.45$) & $3.12$ & $1.76$ \\
    \texttt{o1/0} & $1.62$ ($1.78$) & $2.10$ & $3.27$ ($0.28$) & $4.32$ & $3.33$ \\
    \midrule
    mean & $1.40\,{\pm}\,0.29$ & $1.60\,{\pm}\,0.51$ & $2.11\,{\pm}\,1.03$ & $3.81\,{\pm}\,0.62$ & $2.14\,{\pm}\,1.05$ \\
    \bottomrule
  \end{tabular}
\end{table}

\begin{figure}[htbp]
  \centering
  \includegraphics[width=0.9\linewidth]{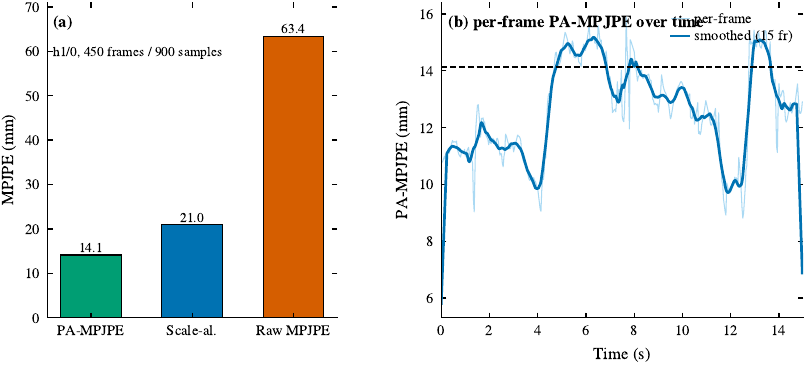}
  \caption{V-HPOT hand accuracy on H2O \texttt{h1/0}. (a) Summary bars: scale-invariant PA-MPJPE ($14.13\,\mathrm{mm}$), scale-aligned MPJPE ($22.0\,\mathrm{mm}$), and absolute (metric) MPJPE ($63.7\,\mathrm{mm}$) over $900$ hand samples. (b) Per-frame PA-MPJPE over time (light) with a $15$-frame smoothed curve and the sequence mean (dashed). The per-frame PA-MPJPE is broadly stable over the sequence. 20 sequences: $14.53{\pm}1.29$ PA-MPJPE, $26.46{\pm}4.40$ scale-aligned (mean$\pm$std).}
  \label{fig:accuracy}
\end{figure}

\paragraph{Hand pose.}
Running the pretrained V-HPOT \emph{frozen} (no TTO, matching the release) on all 20 H2O sequences ($12{,}032$ frames, $21{,}876$ samples) gives a PA-MPJPE of $14.53\,{\pm}\,1.29\,\mathrm{mm}$ and a scale-aligned MPJPE of $26.46\,{\pm}\,4.40\,\mathrm{mm}$ against GT; on \texttt{h1/0} the PA-MPJPE is $14.13\,\mathrm{mm}$ and the per-frame curve is stable (Figure~\ref{fig:accuracy}). Because the anchor is metric we also report the absolute (unaligned) MPJPE: $63.7\,\mathrm{mm}$ on \texttt{h1/0} and $119.1\,{\pm}\,31.3\,\mathrm{mm}$ over the 20 sequences (pooled $117.9$), with a Procrustes scale of $0.73$ ($0.71\,{\pm}\,0.07$)---the anchored skeleton is systematically about $37\%$ too large, the cost of the box-median anchor. The aligned metrics characterize hand configuration, while absolute MPJPE exposes the scale and localization errors that remain after anchoring. All results use the pretrained checkpoint without target-domain training.

\paragraph{Object and body streams.}
The detection/tracking/segmentation stream is built from LocateAnything-3B~\cite{locateanything2025}, GMC-BoT-SORT~\cite{aharon2022botsort} and SAM~2.1 (Hiera-L)~\cite{ravi2024sam2}. H2O provides no instance masks for our categories, so we do not re-benchmark it on H2O and instead rely on the checkpoints' published benchmarks (LocateAnything-3B on COCO/LVIS in its model card; we use BoT-SORT with its ReID head disabled; SAM~2.1's $79.5$ J\&F on SA-V test). On \texttt{h1/0} the composed stream yields nine object tracks, seven persisting more than $30$ frames (median coverage $0.75$); the three named instances (bottle, cup, box) are detected in $450$, $450$ and $420$ frames, while open-vocabulary ``object'' tracks are more fragmented ($0.32$--$0.75$), the expected weak point of zero-shot detection on small or occluded instances. Whole-body pose is qualitative and domain-separate (Section~\ref{sec:whole_body_estimation}).

\subsection{Cross-Modal Consistency Validation}
\label{sec:consistency_eval}
We next examine agreement across the registered annotation streams. Figure~\ref{fig:consistency} illustrates the hand--depth residual on a representative sequence, and Table~\ref{tab:consistency} summarizes the 20-sequence hand study and the object smoothness case study. The constant-scale control isolates the effect of per-hand anchoring beyond basic unit conversion.

\begin{table}[htbp]
  \centering
  \caption{Cross-modal consistency on H2O. Hand--depth residual: pooled median over 20 sequences ($429{,}896$ joints), with the per-sequence mean$\pm$std of the medians below it. Object-trajectory smoothness: median on \texttt{h1/0}. ``Naive'' applies no unit conversion; ``Const-scale'' applies one global scale; ``Per-hand'' is our anchoring (Eq.~\eqref{eq:scale_correction}); ``GT'' reprojects the dataset's ground-truth hands onto the same depth map as a GT reference residual. Aggregation differs by row: the residual is a pooled median over all joint samples, with the distribution of per-sequence medians underneath, and the smoothness is a median over object-frame samples at H2O's fixed $30$\,fps. Lower is better.}
  \label{tab:consistency}
  \begin{tabular}{lcccc}
    \toprule
    Measure & Naive & Const-scale & Per-hand & GT ref. \\
    \midrule
    Hand--depth residual, median (20 seq.) & unusable$^{a}$ & $101.1\,\mathrm{mm}$ & $84.6\,\mathrm{mm}$ & $20.5\,\mathrm{mm}$ \\
    \quad per-sequence median & --- & $111.7\,{\pm}\,43.9$ & $86.4\,{\pm}\,16.9$ & $22.2\,{\pm}\,6.0$ \\
    Object-trajectory smoothness, median (h1/0) & --- & --- & $3.67\,\mathrm{mm/frame}^2$ & --- \\
    \bottomrule
  \end{tabular}

  \vspace{2pt}
  {\footnotesize $^{a}$V-HPOT's raw output is in virtual-camera units (residual $\approx7.4\times10^{5}\,\mathrm{mm}$); this is a unit mismatch, not a meaningful distance.}
\end{table}

\paragraph{Hand--depth residual.}
Across 20 sequences, the pooled median hand--depth residual is $84.6\,\mathrm{mm}$ (per-sequence $86.4\pm16.9$, range $64.2$--$119.7$). The distribution has a heavy tail: $p_{75}=136.1$, $p_{90}=185.8$, and $p_{99}=344.3\,\mathrm{mm}$, with $42.0\%$ of joints exceeding $100\,\mathrm{mm}$, consistent with the occlusion failures discussed in Section~\ref{sec:failure}. Reprojected ground-truth hands yield a median of $20.5\,\mathrm{mm}$, approximately one quarter of the predicted residual. This reference is not a strict lower bound: annotated joints lie inside the hand, whereas depth measures the visible surface, and occlusion, projection error, and invalid depth affect both measurements.

A \emph{wrist-excluded} diagnostic anchors the hand using wrist-pixel depth and evaluates the remaining 20 joints. It yields $64.6\,\mathrm{mm}$ (per-sequence $64.8\,{\pm}\,11.2$), compared with $90.0\,\mathrm{mm}$ for box anchoring on the same joints. This comparison characterizes sensitivity to the anchoring region, but it does not remove dependence on the calibration depth: at $320{\times}176$ resolution, different joints may sample the same nearest-neighbor cell. Moreover, a camera-frame residual cannot detect errors confined to the camera-to-world transformation.

The RGB-only diagnostic shows the same ordering between per-hand and constant-scale anchoring: $9.8\%$ versus $12.7\%$ on \texttt{h1/0}, and $11.0\%$ versus $15.5\%$ on \texttt{o1/0} (Table~\ref{tab:consistency_noinj}). These residuals are normalized by wrist depth because MonST3R's RGB-only depth is defined only up to scale.

\begin{table}[htbp]
  \centering
  \caption{Sensor-depth-free (RGB-only) hand--depth residual on the two sequences that have V-HPOT tracks. MonST3R's predicted depth is defined only up to a scale, so the residual is \emph{normalized by the wrist depth} (dimensionless); ``per-hand'' uses the per-frame box-median ratio and ``const'' a single per-sequence ratio. These values must not be compared with the millimeter residuals of Table~\ref{tab:consistency}.}
  \label{tab:consistency_noinj}
  \small
  \begin{tabular}{@{}lccc@{}}
    \toprule
    Sequence & joints & per-hand & const-scale \\
    \midrule
    \texttt{h1/0} & $18{,}900$ & $9.8\%$ & $12.7\%$ \\
    \texttt{o1/0} & $18{,}144$ & $11.0\%$ & $15.5\%$ \\
    \bottomrule
  \end{tabular}
\end{table}

\paragraph{Value of per-hand anchoring.}
On \texttt{h1/0}, virtual-camera conversion already yields a near-metric scale (fitted global scale $\approx1.002$), and per-hand anchoring gives nearly the same residual as a single global scale ($66.6$ versus $66.5\,\mathrm{mm}$). Across 20 sequences, however, the pooled median decreases from $101.1$ to $84.6\,\mathrm{mm}$, a $16.3\%$ reduction. The corresponding per-sequence medians are $111.7\pm43.9$ and $86.4\pm16.9\,\mathrm{mm}$: anchoring reduces their mean by $23\%$ and improves 17 of 20 sequences.

To examine sensitivity to depth-scale drift, we apply a smooth per-frame factor $g_t$ to the depth map, $D'_t = g_t D_t$. Before temporal smoothing, whole-hand anchoring propagates this perturbation as $J'_t = g_t J_t$, leaving pixel projections unchanged and scaling the residual as $r'_t = g_t r_t$. Depth normalization can therefore cancel the common factor even when the metric scale is wrong; temporal smoothing further couples neighboring frames. The experiment measures how the two estimators respond to a shared perturbation, rather than whether they recover true scale. With a $30\%$ perturbation amplitude on ten sequences, the constant-scale residual increases by $19\%$ ($105.5 \to 125.5\,\mathrm{mm}$), while the per-hand residual changes little ($86.9 \to 86.4\,\mathrm{mm}$).

We also restrict the data used to estimate the global scale. Fitting it on the first half of each sequence yields a pooled median of $107.9\,\mathrm{mm}$ (per-sequence $123.7\,{\pm}\,62.8$); fitting on ten sequences and applying it to the other ten yields $106.9\,\mathrm{mm}$. Both exceed the frame-local anchor's residual ($85.0\,\mathrm{mm}$; per-sequence $87.0\,{\pm}\,17.1$). These controls compare estimation ranges: the per-hand anchor still uses the current frame's depth, so the holdouts do not provide an independent validation of that anchor.

\paragraph{Object-trajectory smoothness.}
On \texttt{h1/0}, median object-trajectory acceleration is $3.67\,\mathrm{mm/frame}^2$, equivalent to $3.3\times10^{3}\,\mathrm{mm/s^2}$ at $30$\,fps. The static bottle and cup yield $4.68$ and $1.29\,\mathrm{mm/frame}^2$, respectively. The manipulated box moves $142\,\mathrm{mm}$ while maintaining a median acceleration of $2.18\,\mathrm{mm/frame}^2$. These values describe temporal stability, subject to the world-frame scale limitations in Table~\ref{tab:provenance}. A low acceleration may also result from a stable tracking bias or excessive smoothing, so it does not establish object accuracy. Evaluation against object-pose ground truth remains future work.

\paragraph{GT-assisted single-object trajectory case study.}
The single annotated object in \texttt{h1/0} is present in all $450$ frames. We fix its track association using only the \emph{first frame}: the ground-truth object center is projected with the camera intrinsics, and the track whose box contains that projection is selected. Of the five tracks present at frame $0$, exactly one satisfies this condition---the \texttt{box} track, with $420$ common frames. This is a GT-assisted association, but it avoids selecting tracks independently by ground-truth proximity at every frame.

In camera coordinates, the absolute centroid-to-center error has a median of $54.4\,\mathrm{mm}$ and a mean of $57.7\,\mathrm{mm}$. The error is dominated by the offset between the back-projected visible surface and the annotated object origin, so it is not an object-pose metric. Estimating a fixed offset on the first half of the frames and evaluating on the second yields a held-out \emph{trajectory-shape} error of $12.7\,\mathrm{mm}$ (mean $14.0$; the whole-sequence-offset result is $11.9$). Total displacement is $105.9\,\mathrm{mm}$, compared with $111.0\,\mathrm{mm}$ for ground truth. The result supports motion consistency for this object and sequence. Unlike world-frame smoothness, this camera-frame comparison does not use predicted poses (Table~\ref{tab:provenance}); ground-truth-free association and multi-object evaluation remain open tasks.

\subsection{Cost Analysis}
\label{sec:cost}
MonST3R dominates the computational cost. On a single NVIDIA GPU with $24\,\mathrm{GB}$ of memory, it processes a $351$-frame sequence in approximately $10$ minutes and peaks near $23\,\mathrm{GB}$ of GPU memory. V-HPOT processes $450$ frames in approximately $14$ seconds; object detection, tracking, segmentation, and bookkeeping add a few minutes. End-to-end processing takes roughly $10$--$15$ minutes per sequence, with observed per-frame times of approximately $1.7$--$2.6$ seconds across the $351$- and $450$-frame runs. These timings come from separate runs and are not pooled into a single throughput estimate. The pipeline requires no per-frame human supervision. Reducing the MonST3R resolution ($320{\rightarrow}256$) or iteration count could lower runtime, although the accuracy trade-off has not been evaluated. We do not estimate manual time savings because a comparable annotation rate for this combination of 3D modalities is unavailable.

\subsection{Failure Analysis}
\label{sec:failure}
We identify four failure modes in the H2O runs and the additional compatibility checks.
\emph{(i) Hand--depth disagreement under occlusion.} Across the 20 sequences, $42.2\pm9.4\%$ of per-joint samples exhibit a hand--depth residual above $100\,\mathrm{mm}$ (h1/0: $29.1\%$; worst sequence: $61.2\%$). These concentrate where the hand is partially occluded by an object so that the sampled sensor depth falls on the occluder rather than the hand, or where the hand is in fast motion and the sensor depth is noisy or missing. Scale-invariant PA-MPJPE can remain low when the relative joint configuration is preserved despite errors in absolute scale or position.
\emph{(ii) Scale drift without sensor depth.} On recordings without sensor depth, such as ML-EgoDex, MonST3R's scale ambiguity propagates to the depth map and the anchored hands. The response analysis in Section~\ref{sec:consistency_eval} shows that local anchoring follows changes in depth scale. A stable normalized residual can therefore coexist with incorrect metric geometry.
\emph{(iii) Reflective and specular surfaces.} Specular highlights on bottles and metallic objects produce depth spikes that bias both object back-projection and the object-trajectory smoothness score.
\emph{(iv) Missed hands under extreme occlusion.} V-HPOT predicts a per-hand presence score; on \texttt{h1/0} both hands are detected in all $450$ frames ($900$ of $900$ slots), and the rare misses in our broader trials correspond to the hand being fully outside the field of view or completely hidden behind an object. The presence score flags these cases, and the corresponding predictions are omitted.

\subsection{Qualitative Results}
\begin{figure}[htbp]
  \centering
  \includegraphics[width=0.7\linewidth]{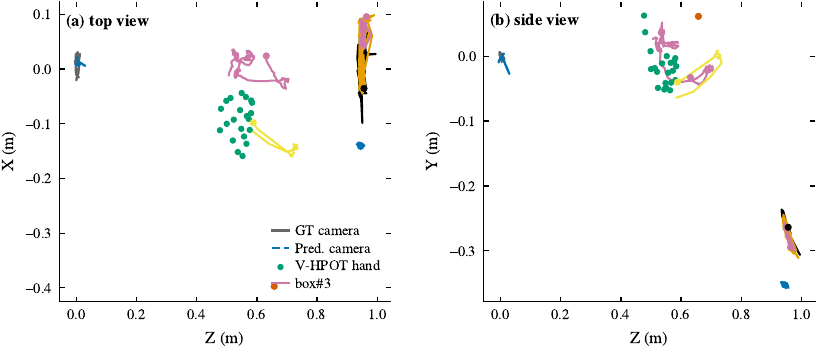}
  \caption{Fused 3D scene on H2O (frame $225$), in the first-frame camera frame. Gray: ground-truth camera trajectory; blue: predicted camera trajectory ($\mathrm{Sim}(3)$-aligned); green: the V-HPOT hand skeleton after metric-scale anchoring; colored lines: 3D trajectories of the tracked objects lifted through the depth map and the camera pose (start points marked). The independently estimated hand, object, and camera streams are drawn in one coordinate system for inspection.}
  \label{fig:scene3d}
\end{figure}

Figures~\ref{fig:pipeline_output} and~\ref{fig:tracking_results} illustrate the integrated viewer and the object tracking and segmentation outputs. The viewer places predicted hands, object points, camera trajectories, and ground-truth references under distinct entity paths, allowing frame-by-frame inspection of agreement and failure. Figure~\ref{fig:scene3d} shows a representative 3D scene with anchored hands, lifted object trajectories, and camera trajectories. This visualization supports qualitative inspection subject to the scale assumptions in Section~\ref{sec:consistency}; the predicted camera trajectory shown in the figure is $\mathrm{Sim}(3)$-aligned to ground truth. Structured captions provide an additional qualitative downstream artifact (Section~\ref{sec:captioning}).

\subsection{Limitations}
\label{sec:limitations}
The evaluation establishes component accuracy and several consistency diagnostics, with the following limitations.
\emph{Calibration dependence and metric scale.} The main hand and object studies use sensor depth. Because hand anchoring and residual evaluation share that depth, the residual cannot independently establish registration accuracy. Wrist-excluded and holdout analyses characterize this dependence without eliminating it. RGB-only depth remains scale-ambiguous, and the implemented sensor-depth initialization does not preserve depth during window optimization. In particular, combining metric sensor-depth points with camera translations at a different scale limits the interpretation of world-frame object trajectories. Independent geometric validation is needed to establish metric accuracy.
\emph{Coverage.} Quantitative evaluation is limited to H2O and one subject. Hand results cover 20 sequences across three action groups, camera ATE covers three sequences, and the RGB-only hand diagnostic covers two. Object results comprise a single-sequence stability analysis and a GT-assisted single-object trajectory study; the 20-sequence hand runs contain no object detections. Detection and segmentation lack mask-based evaluation, and whole-body pose and captioning are qualitative demonstrations. Broader subject, scene, and modality coverage is needed to assess generalization.
\emph{Statistics and baselines.} Pooled medians, means of per-sequence statistics, and per-frame curves summarize different distributions and should not be used interchangeably. Our controls are internal; no competing end-to-end auto-labeling system is evaluated. The cited HaMeR result uses a different protocol and provides context only.

\section{Conclusion}
\label{sec:conclusion}
We presented EgoLabel, a zero-shot pipeline that integrates pretrained geometry, object, and hand models through explicit coordinate and scale conversions. Its contribution is an auditable registration procedure accompanied by diagnostics that expose cross-modal disagreement and distinguish relative consistency from metric accuracy. On 20 sensor-assisted H2O sequences, per-hand anchoring reduces the mean of per-sequence median hand--depth residuals by $23\%$ relative to a single global scale. The pooled residual remains $84.6\,\mathrm{mm}$ against a $20.5\,\mathrm{mm}$ ground-truth reference, and the gap between PA-MPJPE ($14.53\,{\pm}\,1.29\,\mathrm{mm}$) and absolute MPJPE ($119.1\,{\pm}\,31.3\,\mathrm{mm}$) highlights the remaining scale error. On three RGB-only sequences, camera ATE is $1.40\,{\pm}\,0.29\,\mathrm{cm}$ with $\mathrm{Sim}(3)$ alignment and $1.60\,{\pm}\,0.51\,\mathrm{cm}$ with $\mathrm{SE}(3)$ alignment. The sensor-depth initialization ablation further shows that implementation-level assumptions about fixed depth must be checked explicitly. Future work will prioritize independent validation of hand anchoring, consistent metric geometry, broader RGB-only evaluation, and quantitative assessment of object and body annotations. Code, configurations, and evaluation scripts will be released upon publication; a one-command reproduction pipeline is in preparation.

{\small
\bibliographystyle{unsrt}  
\bibliography{references}
}

\end{document}